          [2001/04/25 1.1 (PWD)]
\documentclass[a4paper,twocolumn]{esapub2005} 
\usepackage{times}
\usepackage{natbib}
\usepackage{url}
\usepackage{graphicx}
\usepackage{multirow}
\usepackage{amsmath}
\usepackage{amssymb}
\usepackage{booktabs}

\title{Bridging the Synthetic–Real Gap: Supervised Domain Adaptation for Robust Spacecraft 6-DoF Pose Estimation}
\author{Inder Pal Singh}
\author{Nidhal Eddine Chenni}
\author{Abd El Rahman Shabayek}
\author{Arunkumar Rathinam}
\author{Djamila Aouada}
\affil{SnT, University of Luxembourg, Luxembourg\\
\{inder.singh, nidhal.chenni, abdelrahman.shabayek, arunkumar.rathinam, djamila.aouada\}@uni.lu}

\begin{document}


\maketitle
\begin{abstract}
Spacecraft Pose Estimation (SPE) is a fundamental capability for autonomous space operations such as rendezvous, docking, and in-orbit servicing. Hybrid pipelines that combine object detection, keypoint regression, and Perspective-$n$-Point (PnP) solvers have recently achieved strong results on synthetic datasets, yet their performance deteriorates sharply on real or lab-generated imagery due to the persistent synthetic-to-real domain gap. Existing unsupervised domain adaptation approaches aim to mitigate this issue but often underperform when a modest number of labeled target samples are available. In this work, we propose the first Supervised Domain Adaptation (SDA) framework tailored for SPE keypoint regression. Building on the Learning Invariant Representation and Risk (LIRR) paradigm, our method jointly optimizes domain-invariant representations and task-specific risk using both labeled synthetic and limited labeled real data, thereby reducing generalization error under domain shift. Extensive experiments on the SPEED+ benchmark demonstrate that our approach consistently outperforms source-only, fine-tuning, and oracle baselines. Notably, with only 5\% labeled target data, our method matches or surpasses oracle performance trained on larger fractions of labeled data. The framework is lightweight, backbone-agnostic, and computationally efficient, offering a practical pathway toward robust and deployable spacecraft pose estimation in real-world space environments.
\end{abstract}

\section{Introduction}
The determination of the position and orientation constituting six-degree-of-freedom (6-DoF) pose of a target spacecraft, commonly referred to as spacecraft pose estimation, is a key enabler for autonomous on-orbit operations such as rendezvous, docking, inspection, and servicing \cite{tarabini2013proba, d2012spaceborne}. The monocular vision-based 6-DoF pose estimation involves deducing the rotation and translation data of a target spacecraft from 2D images captured from the chaser spacecraft. Accurate estimation of the 6-DoF pose from 2D images is particularly challenging in the space environment due to illumination changes, specular reflections, limited texture, and significant variations in the apparent size of the target caused by changes in range during approach \cite{hu2021wide}.

\begin{figure*}[t]
	\centering
	\includegraphics[width=.8\linewidth]{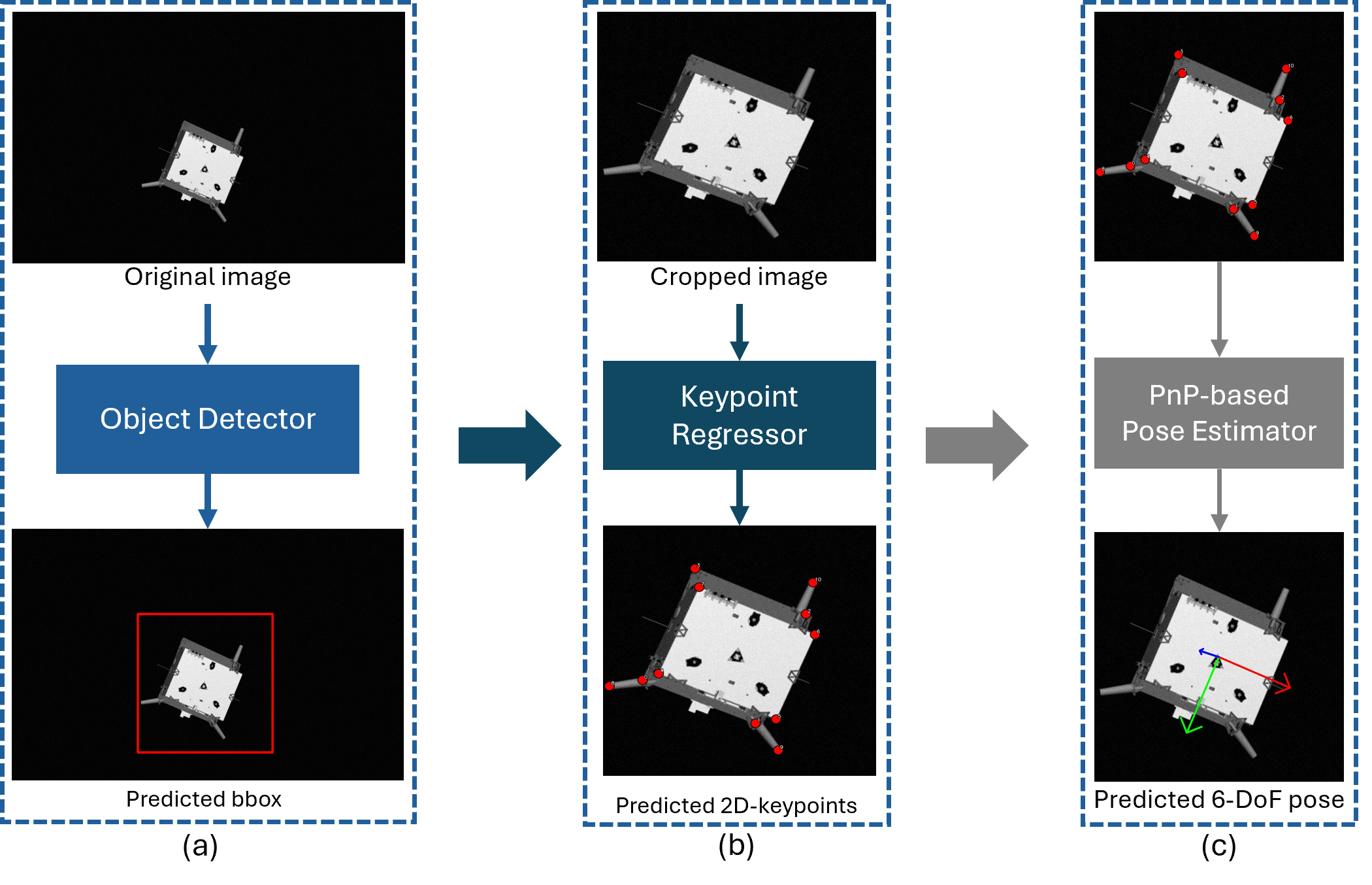}
	\caption{Working pipeline for hybrid modular approaches.}
	\label{fig:hybrid}
\end{figure*}

Early spacecraft pose estimation methods relied primarily on geometric computer vision techniques such as edge matching, template alignment, and photogrammetry-based measurements \cite{petit2012vision}. Although effective in controlled conditions, these methods are sensitive to noise and environmental variations, which limit their robustness in real on-orbit imagery.  
With the rise of deep learning (DL), fully end-to-end networks have been explored, mapping raw images directly to pose parameters \cite{park2019towards, park2024robust}. Such approaches can implicitly learn complex visual cues, but often require vast amounts of labelled data and can be less interpretable or adaptable to new spacecraft geometries.

Alternate approaches to end-to-end approaches are hybrid modular approaches to estimate spacecraft poses (Figure~\ref{fig:hybrid}). These methods combine data-driven feature extraction with geometric model-based solvers, exploiting the strengths of both paradigms \cite{pauly2023survey, ye2023filterformerpose}. In a typical pipeline, the process begins with \emph{spacecraft localization}, where a deep learning (DL) object detection model predicts a bounding box enclosing the target spacecraft (see Figure~\ref{fig:hybrid}(a)). This stage mitigates the effects of background clutter and scale variation by extracting a region of interest (ROI) for subsequent processing. 

The next stage, \emph{keypoint regression}, employs a DL-based regression network to infer the two-dimensional image coordinates of a predefined set of three-dimensional (3D) keypoints on the spacecraft (see Figure~\ref{fig:hybrid} (b)). These keypoints are typically chosen to correspond to geometrically distinctive features such as the vertices of structures or distinct locations on the spacecraft. The 3D coordinates are obtained from an a priori spacecraft 3D model or be reconstructed from imagery via structure-from-motion techniques.

Finally, as shown in Figure~\ref{fig:hybrid} (c), the \emph{pose computation} stage establishes correspondences between the regressed 2D keypoints and the known 3D keypoints, solving the Perspective-$n$-Point (PnP) problem to recover the translation vector and rotation quaternion \cite{lepetit2009ep}. To improve accuracy, the initial PnP solution can be refined by minimizing a geometric reprojection error through optimization. 

Several studies \cite{park2024robust,rathinam2024spades} have shown that this hybrid modular strategy achieves better accuracy over direct approaches in benchmark datasets such as SPEED \cite{speedC} and SPEED+ \cite{speedplusC} while preserving modularity, enabling individual components to be upgraded without retraining the entire pipeline. This modularity is especially advantageous in operational contexts where mission-specific spacecraft geometries or updated sensor models must be accommodated without complete end-to-end retraining.

Despite their demonstrated effectiveness, current hybrid modular methods are significantly affected by the \emph{domain gap} that arises from the discrepancies between synthetic and real imagery. Due to the scarcity and cost of acquiring annotated spacecraft datasets, most existing models are trained predominantly on synthetic images, which often fail to capture the full complexity of real on-orbit conditions, especially physical interaction of different materials under different lighting conditions. As a result, models that perform well on synthetic benchmarks may exhibit substantial performance degradation when deployed on real mission data \cite{pauly2023survey}. Domain adaptation techniques have been proposed to mitigate this gap, with most focussing on unsupervised adaptation scenarios where target domain labels are unavailable \cite{perez2022spacecraft, perez2023spacecraft}. However, in many practical situations, a limited set of labelled target-domain images can be obtained, offering the opportunity to further improve performance beyond what unsupervised methods can achieve. To this end, we aim to exploit this setting by jointly learning domain-invariant representations and minimizing task-specific risk, thereby leveraging both synthetic and limited labelled real data to enhance robustness and accuracy in spacecraft pose estimation.

The main contributions of this work are as follows. First, we investigate the impact of the synthetic-to-real domain gap on hybrid modular spacecraft pose estimation pipelines, highlighting performance degradation when models are trained solely on synthetic data. Second, we propose a domain adaptation framework that exploits limited labelled target-domain data, a scenario that is often realistic in space applications. The approach jointly learns domain-invariant feature representations and minimizes task-specific risk, enabling the model to leverage both synthetic and limited real data effectively. Finally, we evaluate the proposed method on benchmark datasets, demonstrating improved robustness and accuracy compared to state-of-the-art unsupervised adaptation and baseline hybrid modular methods.

\section{Problem Formulation}
This section presents mathematical formulation of the 6-DoF SPE problem. Later, the formulation extended to the SDA setting, where a model trained predominantly on synthetic (source) data must adapt to scarcely labelled real (target) data. Finally, the problem setup to derive finite-sample generalization bounds under the LIRR~\cite{lirr} strategy, providing a theoretical basis to understand performance limits with this setup.

\subsection{6-DoF Pose Estimation via Hybrid Modular Approach}
In the hybrid modular framework, the 6-DoF SPE problem is decomposed into three sequential stages: (i) prediction of the bounding box for spacecraft localization, (ii) 2D keypoint regression, and (iii) Perspective-$n$-Point (PnP) pose computation. This decomposition enables independent optimization of each module, leveraging both deep learning and geometric constraints.

\paragraph{Bounding Box Prediction}
Let $\mathcal{I}$ denote an input image captured by the monocular camera. A deep learning-based object detection model predicts an axis-aligned bounding box $\mathbf{b} = (u_{\min}, v_{\min}, u_{\max}, v_{\max})$ that encloses the target spacecraft in the image plane. This prediction can be formulated as:
\begin{equation}
    \hat{\mathbf{b}} = \arg\max_{\mathbf{b}} \ P_{\theta}(\mathbf{b} \mid \mathcal{I}),
\end{equation}
where $P_{\theta}$ is the detection model parameterized by $\theta$. The predicted bounding box defines a region of interest (ROI) that is cropped and passed to the subsequent keypoint regression stage. Since this work focusses exclusively on the keypoint regression stage, the bounding box coordinates are obtained from ground-truth annotations rather than being predicted by an object detection model.

\paragraph{2D Keypoint Regression}
Let $\mathcal{P} = \{\mathbf{P}_i \in \mathbb{R}^3 \mid i=1, \dots, N\}$ denote the set of $N$ predefined three-dimensional (3D) keypoints on the target spacecraft, expressed in the spacecraft body-fixed frame. Given the cropped ROI $\mathcal{I}_{\mathrm{ROI}}$, a regression network $f_{\phi}$ predicts the 2D image coordinates $\hat{\mathbf{p}}_i \in \mathbb{R}^2$ of each keypoint:
\begin{equation}
    \hat{\mathbf{p}}_i = f_{\phi}(\mathcal{I}_{\mathrm{ROI}}), \quad i = 1, \dots, N.
\end{equation}
In general, the network outputs heatmaps that represent the probability distribution of each keypoint location, from which $\hat{\mathbf{p}}_i$ is obtained via an argmax or soft-argmax operation.

\paragraph{PnP-Based Pose Computation}
The goal of the final stage is to estimate the rigid-body transformation $\mathbf{T} \in SE(3)$ mapping the spacecraft body frame to the camera frame:
\begin{equation}
    \mathbf{T} = \begin{bmatrix} \mathbf{R} & \mathbf{t} \\ \mathbf{0}^\top & 1 \end{bmatrix},
\end{equation}
where $\mathbf{R} \in SO(3)$ is the rotation matrix (parameterized by a unit quaternion $\mathbf{q}$) and $\mathbf{t} \in \mathbb{R}^3$ is the translation vector.  

Let $\mathbf{K} \in \mathbb{R}^{3\times3}$ be the intrinsic calibration matrix of the camera, and $\tilde{\mathbf{P}}_i \in \mathbb{R}^4$ and $\tilde{\mathbf{p}}_i \in \mathbb{R}^3$ be the homogeneous coordinates of the $i$-th 3D keypoint and its 2D image projection, respectively. The projection model is:
\begin{equation}
    s_i \tilde{\mathbf{p}}_i = \mathbf{K} \left[ \mathbf{R} \ \mathbf{t} \right] \tilde{\mathbf{P}}_i,
\end{equation}
where $s_i$ is a scale factor.  

Given $\{\hat{\mathbf{p}}_i\}$ from the regression stage and $\{\mathbf{P}_i\}$ from the known spacecraft model, the 6-DoF pose estimation problem becomes:
\begin{equation}
    \hat{\mathbf{R}}, \hat{\mathbf{t}} = \arg\min_{\mathbf{R}, \mathbf{t}} \sum_{i=1}^N \left\| \mathbf{p}_i - \pi(\mathbf{R}\mathbf{P}_i + \mathbf{t}) \right\|_2^2,
\end{equation}
where $\pi(\cdot)$ is the projection function defined by $\mathbf{K}$. This corresponds to the classical PnP problem \cite{lepetit2009ep}, which can be solved using methods such as EP$n$P, followed by non-linear refinement to minimize the reprojection error.

\subsection{Supervised Domain Adaptation (SDA)}
Let $\mathcal{X}$ and $\mathcal{Y}$ denote the input and output spaces, respectively, and let $\mathcal{Z}$ be the representation space such that a feature extractor $g: \mathcal{X} \rightarrow \mathcal{Z}$ maps the input samples to their latent representations. We denote by $X, Y, Z$ the random variables that take the values in $\mathcal{X}, \mathcal{Y}, \mathcal{Z}$, respectively. In a domain adaptation setting, $\mathcal{D}_S$ and $\mathcal{D}_T$ represent the source and target domains, and $D \in \{S, T\}$ is a categorical variable indicating the domain index.

In SDA, we have access to $n$ labelled samples from the source domain and $m$ labelled samples from the target domain, with $n \gg m$. The objective is to leverage both the labelled source and the limited labelled target data to train a hypothesis $h$ that generalizes well to $\mathcal{D}_T$.

Formally, the \textit{risk} of a hypothesis $h$ with respect to the true labelling function $f$ in the domain $\mathcal{D}_S$ is defined as:
\[
\varepsilon_S(h) := \mathbb{E}_{(X,Y) \sim \mathcal{D}_S} \left[ \ell(h(X), Y) \right]
\]
where $\ell(\cdot)$ is a loss function (for example, MSE for regression) and$\hat{\varepsilon}_S(h)$ denotes the empirical risk of $h$ in the source domain. Similarly, $\varepsilon_T(h)$ and $\hat{\varepsilon}_T(h)$ represent the true and empirical risks in the target domain, respectively.

The core challenge in domain adaptation is to ensure that a small empirical source error $\hat{\varepsilon}_S(h)$ implies a small target error $\varepsilon_T(h)$, which is generally not possible when the source and target distributions differ significantly. 



\subsection{Generalization Bounds}
The proposed approach adapts the finite-sample generalization bounds from the original LIRR strategy~\cite{lirr} in the context of SDA for regression tasks.

Let $\mathcal{H}$ be a hypothesis set with functions $h : \mathcal{Z} \rightarrow [0, 1]$ and $Pdim(\mathcal{H}) = d$. Let $\hat{D}_S$ and $\hat{D}_T$) denote the empirical distributions induced by samples from the source domain $\mathcal{D}_S$ and target domain $\mathcal{D}_T$, respectively. The SDA setup has access to $n$ labelled samples from the source domain $S$ and $m$ labelled samples from the target domain $T$, with $n \gg m$ in typical scenarios.

We define the hypothesis set $\tilde{\mathcal{H}} := \{\mathbb{I}[|h(x) - h'(x)| > t] : h, h' \in \mathcal{H}, 0 \le t \le 1\}$. For $0 < \delta < 1$, with probability at least $1 - \delta$ over the sampling of $n$ points from $S$ and $m$ points from $T$, the following bound holds for all $h \in \mathcal{H}$:

\begin{align}
\varepsilon_T(h) &\le \frac{m}{n+m} \hat{\varepsilon}_T(h) + \frac{n}{n+m} \hat{\varepsilon}_S(h)+ \nonumber \\
&\quad \frac{n}{n+m} \Big\{ d_{\tilde{\mathcal{H}}}(\hat{D}_S(Z), \hat{D}_T(Z))+ \nonumber \\
&\quad \min \Big[ \mathbb{E}_S \| f_S(Z) - f_T(Z) \|,\nonumber \\
&\quad \quad \mathbb{E}_T \| f_S(Z) - f_T(Z) \| \Big] \Big\}+ \nonumber \\
&\quad \frac{n}{n+m} |n_S + n_T|+ \nonumber \\
& O\left( \sqrt{\left(\frac{1}{m} + \frac{1}{n}\right)\log\frac{1}{\delta} + \frac{d}{n} \log \frac{n}{d} + \frac{d}{m} \log \frac{m}{d}} \right)
\label{eq:bound}
\end{align}

Here, $\varepsilon_T(h)$ denotes the true risk on the target domain, and $\hat{\varepsilon}_S(h)$ and $\hat{\varepsilon}_T(h)$ denote the empirical risks on the source and target domains, respectively. The term $d_{\tilde{\mathcal{H}}}(\hat{D}_S(Z), \hat{D}_T(Z))$ measures the discrepancy between the source and target feature distributions in the representation space $\mathcal{Z}$, and the term $|n_S + n_T|$ captures the amplitude of the noise.

This bound provides a theoretical foundation for SDA in regression by linking the target-domain risk to (i) the empirical risks on the source and target domains, (ii) the distributional discrepancy between domains, (iii) the difference between source and target labeling functions, and (iv) the noise level. In this SDA setup for spacecraft pose estimation, the bound suggests that performance improvements can be achieved by simultaneously minimizing the empirical risk on the target domain and the distributional discrepancy between domains, which is precisely the goal of our proposed LIRR-based adaptation framework.

\section{Methodology}
Motivated by the generalization bounds for regression (\ref{eq:bound}), the proposed approach adapt the LIRR framework~\cite{lirr} to the SDA setting for the spacecraft pose estimation task. The bound decomposes the target-domain error into four main terms:
\begin{enumerate}
    \item Empirical risk on the target domain.
    \item Empirical risk on the source domain.
    \item Discrepancy between the source and target feature distributions.
    \item Discrepancy between the optimal source and target predictors.
\end{enumerate}

In SDA, the first two terms can be minimized directly using labelled source and target data. The third term motivates learning \textit{domain-invariant representations}, while the fourth term motivates learning \textit{invariant risks}. The proposed SDA-LIRR framework jointly optimizes these two objectives.

\subsection{Invariant Representations}

Let $g: \mathcal{X} \rightarrow \mathcal{Z}$ be a feature extractor that maps the inputs $X$ to a latent space $\mathcal{Z}$, and let $D \in \{S, T\}$ be the domain label. The goal of learning invariant representation is to make $Z = g(X)$ statistically independent of the domain $D$. This is equivalent to minimizing the mutual information $I(D; Z)$:
\begin{equation}
    I(D; Z) = 0 \ \ \Rightarrow \ \ p(Z|D = S) = p(Z|D = T)
\end{equation}
which ensures that the third term in the bound vanishes. In practice, this is implemented via adversarial domain classification:
\begin{equation}
    \mathcal{L}_{\text{rep}}(g, C) = \mathbb{E}_{x \sim \mathcal{D}_S} \log C(g(x)) + \mathbb{E}_{x \sim \mathcal{D}_T} \log (1 - C(g(x)))
\end{equation}
where $C$ is a domain classifier trained to distinguish between source and target features, while $g$ is trained to fool $C$, thanks to the incorporation of a Gradient Reversal Layer (GRL).

\subsection{Invariant Risks}
Let \(f : \mathcal{Z} \rightarrow \mathcal{Y}\) denote the task-specific predictor (e.g., spacecraft keypoint regressor) applied to the latent representations produced by a feature extractor \(g : \mathcal{X} \rightarrow \mathcal{Z}\). Invariant risk learning aims to make the conditional distributions \(p(Y \mid Z, D = S)\) and \(p(Y \mid Z, D = T)\) identical, ensuring that the same predictor is optimal for both the source and target domains. In the SDA setting, both the source domain \(\mathcal{D}_S\) and the target domain \(\mathcal{D}_T\) contain labeled samples, with \(|\mathcal{D}_T|\) typically much smaller than \(|\mathcal{D}_S|\).

This objective corresponds to minimizing the conditional mutual information \(I(D; Y \mid Z)\):
\begin{equation}
I(D; Y \mid Z) = 0 \ \Rightarrow \ p(Y \mid Z, D = S) = p(Y \mid Z, D = T).  
\end{equation}

Following \cite{lirr}, \(I(D; Y \mid Z)\) denotes the difference between conditional entropies:
\begin{equation}
I(D; Y \mid Z) = H(Y \mid Z) - H(Y \mid D, Z).  
\label{eq:MI_diff}
\end{equation}
Minimizing Eq.~\eqref{eq:MI_diff} encourages both high predictive accuracy and invariance to the domain label.

Using the variational form of the conditional entropy:
\begin{equation}
H(Y \mid Z) = \inf_{f} \ \mathbb{E}[L(Y; f(Z))],
\end{equation}
where \(L\) denotes the cross-entropy loss, Eq.~\eqref{eq:MI_diff} can be transformed into the minimization of the supervised cross-entropy losses of two predictors:
\begin{itemize}
    \item a \emph{domain-invariant} predictor $(f_i)$, trained using both source and target labeled data,
    \item a \emph{domain-dependent} predictor $(f_d)$, trained with access to both the input and the domain label $D$).
\end{itemize}

The respective optimization problems are:
\begin{align}
\min_{g, f_i} \quad & \mathcal{L}_i = \mathbb{E}_{(x,y) \sim \mathcal{D}_S \cup \mathcal{D}_T} \left[ L\big(y, f_i(g(x))\big) \right], \label{eq:Li} \\
\min_{g, f_d} \quad & \mathcal{L}_d = \mathbb{E}_{(x,y) \sim \mathcal{D}_S \cup \mathcal{D}_T} \left[ L\big(y, f_d(g(x), D)\big) \right]. \label{eq:Ld}
\end{align}

Since \(f_d\) has access to the domain label, we have \(\mathcal{L}_d \le \mathcal{L}_i\). The invariant risk term is then defined as:
\begin{equation}
\mathcal{L}_{\mathrm{risk}}(g, f_i, f_d) = \mathcal{L}_i + \lambda_{\mathrm{risk}}\left( \mathcal{L}_i - \mathcal{L}_d \right),
\label{eq:Lrisk}
\end{equation}
where \(\lambda_{\mathrm{risk}}\) balances the contribution of the discrepancy term.

In practice, minimizing Eq.~\eqref{eq:Lrisk} drives \(f_i\) to match the predictive performance of \(f_d\) without access to domain information, effectively ensuring that the learned representation \(Z = g(X)\) is predictive of \(Y\) but invariant to \(D\). In the SDA setting, the presence of labeled target-domain samples enables \(\mathcal{L}_i\) to be computed directly on both domains, improving adaptation performance compared to unsupervised settings.

\begin{figure*}[!t]
	\centering
	\includegraphics[width=0.9\linewidth]{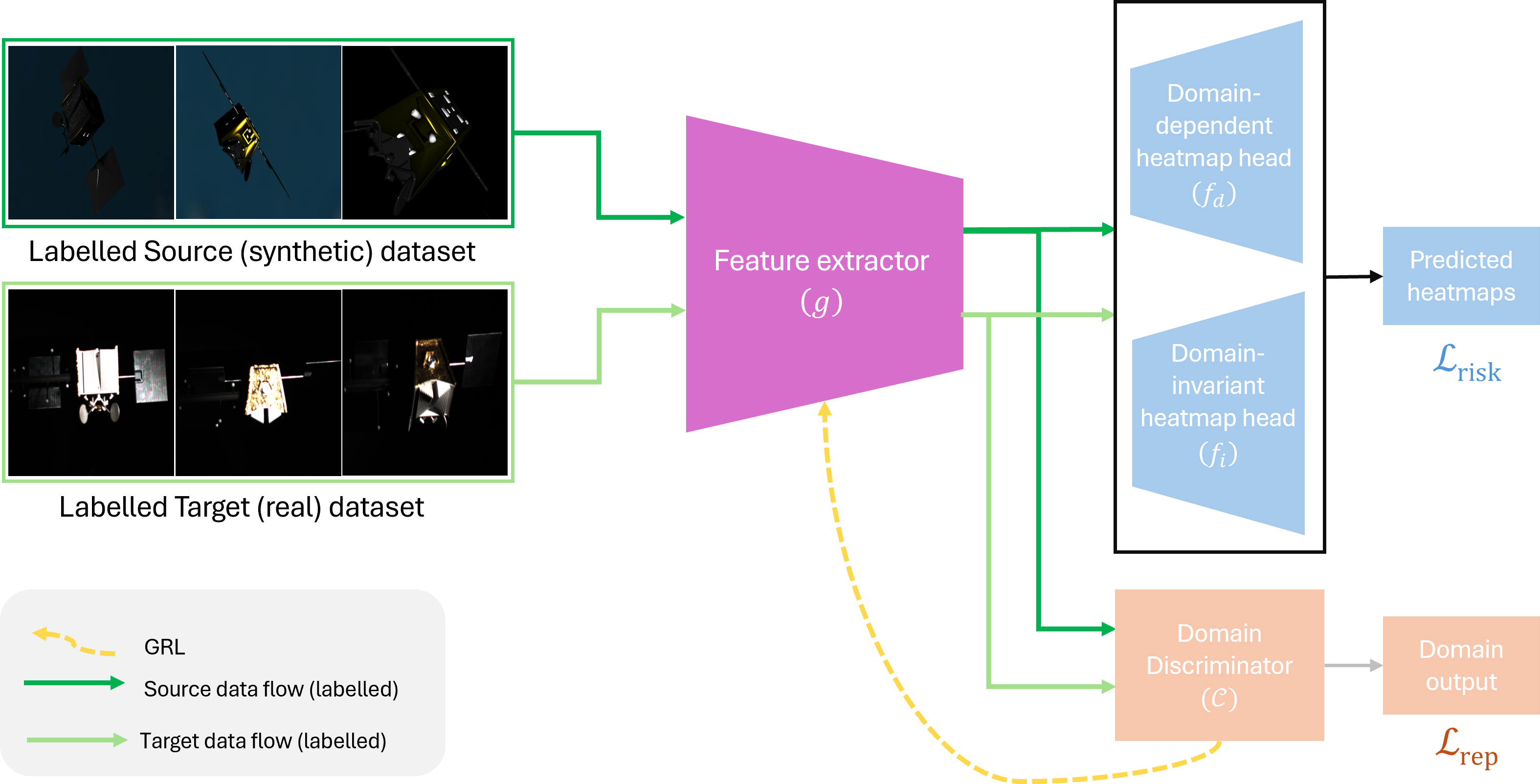}
	\caption{Overview of the proposed LIRR-SDA architecture for spacecraft pose estimation. Synthetic (source) and real (target) labeled images are passed through a shared feature extractor $g$. The extracted features are processed by two heatmap regression heads: a domain-dependent head $f_d$ and a domain-invariant head $f_i$, which together optimize the invariant risk loss $\mathcal{L}_{risk}$. A domain discriminator $\mathcal{C}$, connected via a gradient reversal layer (GRL), enforces feature alignment across domains through the representation loss $\mathcal{L}_{rep}$. This joint learning enables accurate keypoint regression while reducing the synthetic-to-real domain gap.}

	\label{fig:lirr-sda}
\end{figure*}

\subsection{Joint SDA-LIRR Objective for SPE}
In this work, the proposed SDA-LIRR strategy is applied specifically to the task of \emph{keypoint regression} within the hybrid modular spacecraft pose estimation pipeline. As described in Section~2.1, ground truth bounding box is considered for spacecraft localization, thereby isolating the keypoint regression stage as the primary learning target in our domain adaptation framework. This choice removes variability in localization performance and allows us to focus entirely on improving cross-domain generalization for keypoint prediction.

The overall architecture used for this objective is illustrated in Fig.~\ref{fig:lirr-sda}. The images in the source domain (synthetic) and in the target domain (real) are cropped using the GT bounding boxes and passed through a shared \emph{feature extractor} $g$. The resulting feature embeddings are then processed by two parallel heatmap prediction heads: a \emph{domain-dependent} head $f_d$ and a \emph{domain-invariant} head $f_i$. The domain-dependent head captures features specific to each domain, while the domain-invariant head is trained to produce aligned representations across domains, enabling consistent keypoint predictions for both synthetic and real data.

Domain alignment is enforced through a \emph{domain discriminator} $C$, which receives features from $g$ and attempts to predict the domain label. A Gradient Reversal Layer (GRL) ensures that $g$ learns to generate features that confuse the discriminator, thus promoting domain invariance. This process corresponds to the $\mathcal{L}_{\mathrm{rep}}$ term in Eq.~(9), where $\lambda_{\mathrm{rep}}$ balances the alignment of the representation against the invariant risk objective.

The invariant risk loss $\mathcal{L}_{\mathrm{risk}}$ is calculated using the domain-invariant head $f_i$ to ensure that, for matched features, the predicted keypoint heatmaps remain consistent across domains. This joint optimization can be expressed as:
\begin{equation}
    \min_{g, f} \ \max_{C} \ \mathcal{L}_{\mathrm{risk}}(g, f) + \lambda_{\mathrm{rep}} \mathcal{L}_{\mathrm{rep}}(g, C),
\end{equation}
where $\lambda_{\mathrm{rep}}$ controls the trade-off between representation alignment and prediction consistency.

By integrating this SDA-LIRR formulation into the keypoint regression stage, we leverage both the geometric supervision provided by the synthetic domain and the domain adaptation capability from the limited labeled real data. This significantly reduces the synthetic-to-real domain gap in spacecraft pose estimation.

\section{Experiments}

This section presents the evaluation of the proposed approach for SPE in the SDA setting. First, it presents the implementation details, including network architecture, backbone, prediction heads, and training settings. Followed by the details of the datasets used in our experiments and a quantitative performance comparison against the relevant baselines. Qualitative results are presented to illustrate the effectiveness of our method, and finally conclude with an analysis of common failure cases observed in our evaluation.

\subsection{Implementation Details}
\label{subsec:impl_details}
In the implementation, the feature extractor $g$ is based on the MobileNet v2 backbone pre-trained in ImageNet and fine-tuned for the keypoint regression task. The domain-invariant head $f_i$ and domain-dependent head $f_d$ share an identical architecture consisting of interpolation layers that output N heatmaps of size $56\times56$, one for each of the keypoints. The domain discriminator $C$ is implemented as a multi-layer perceptron (MLP). The model is trained using the Adam optimizer with a learning rate of $1e^{-3}$. The input image is cropped to the ground-truth bounding box, expanded by 5\% of its width to ensure context, and resized to $224 \times 224$ before being fed to the network, unless stated otherwise.


\subsection{Dataset overview and configuration}
\label{subsec:datasets}
The proposed approach is evaluated using the SPEED + benchmark dataset for spacecraft domain adaptation. The source domain $(\mathcal{D}_S)$ contains $59,960$ synthetic spacecraft images, while the two target domains (\(\mathcal{D}_T\)), namely Sunlamp and Lightbox, comprise $2,700$ and $6,700$ real laboratory-captured images, respectively. For SDA experiments, $47,966$ labelled source samples and a limited set of labelled target samples were used. Both target domain trainings were considered, each with splits of $250$ and $500$ images, denoted as \(S^{+}_{250}\), \(S^{+}_{500}\) for Sunlamp and \(L^{+}_{250}\), \(L^{+}_{500}\) for Lightbox. Each target dataset has a fixed test set shared across splits, containing $2,200$ images for Sunlamp and $6,200$ for Lightbox.


To assess the effectiveness of the proposed approach, five distinct training scenarios were used and they are listed below.

\begin{enumerate}
    \item \textbf{Source-only}: The model is trained solely on the labelled synthetic source dataset without any domain adaptation. This setting serves as a baseline to quantify the domain gap.

    \item \textbf{Oracle}: The model is trained exclusively on labelled real target-domain data (Sunlamp or Lightbox) using subsets of the available training set (250 and 500 images). This represents an idealized upper bound on performance when full supervision is available in the target domain.

    \item \textbf{Fine-tune}: The model is first trained on the synthetic source dataset and subsequently fine-tuned on subsets of labelled real target-domain images. This setting evaluates the benefit of transfer learning from synthetic to real imagery.

    \item \textbf{SDA}: The model is trained jointly on synthetic and real target-domain labelled images, using a domain discriminator to align feature distributions in an adversarial manner. This implicitly reduces domain discrepancy while leveraging all available labelled data.

    \item \textbf{LIRR-SDA}: Our proposed method extends the LIRR framework~\cite{lirr} to the fully supervised domain adaptation setting. This approach jointly learns domain-invariant representations and explicitly minimizes prediction risk across domains, aiming to maximize cross-domain generalization.
\end{enumerate}

\begin{table*}[t]
\centering
\caption{Keypoint and pose performance for different adaptation algorithms on the SPEED+ Sunlamp and Lightbox domains. 
Best results are highlighted in \textbf{bold}.}
\label{tab:quant_results}
\resizebox{\textwidth}{!}{
\begin{tabular}{|l| l| cc| ccc|}
\toprule
\hline
\textbf{Algorithm} & \textbf{Training set} & \textbf{KError [px]} & \textbf{PCK [\%]} & \textbf{Range error [\% of range]} & \textbf{Attitude error [deg]} & \textbf{ESA Score} \\
\midrule
\hline
\multicolumn{7}{|c|}{\textbf{Sunlamp domain}} \\
\hline
\midrule
SOURCE-ONLY   & SYN                & 193.20 & 11.96 & 0.30 & 76.05 & 1.63 \\
ORACLE        & $S^{+}_{250}$       & 83.11  & 20.78 & 0.13 & 27.41 & 0.61 \\
Fine-tune     & SYN+$S^{+}_{250}$   & 57.40  & 33.00 & 0.14 & 20.97 & 0.51 \\
SDA           & SYN+$S^{+}_{250}$   & 48.37  & 44.40 & 0.13 & 16.96 & 0.43 \\
LIRR-SDA (Ours) & SYN+$S^{+}_{250}$ & 37.17  & 49.78 & 0.09 & 11.58 & 0.29 \\
\midrule
\hline
ORACLE        & $S^{+}_{500}$      & 48.50  & 30.44 & 0.11 & 15.17 & 0.38 \\
Fine-tune     & SYN+$S^{+}_{500}$  & 36.28  & 42.96 & 0.09 & 11.06 & 0.28 \\
SDA           & SYN+$S^{+}_{500}$  & 31.38  & 52.69 & 0.08 & 9.52  & 0.25 \\
LIRR-SDA (Ours) & SYN+$S^{+}_{500}$& \textbf{28.35}  & \textbf{55.56} & \textbf{0.07} & \textbf{7.84}  & \textbf{0.21} \\
\midrule
\hline
\multicolumn{7}{|c|}{\textbf{Lightbox domain}} \\
\midrule
\hline
SOURCE-ONLY   & SYN                & 185.55 & 12.70 & 0.38 & 65.94 & 1.53 \\
ORACLE        & $L^{+}_{250}$     & 98.93  & 17.35 & 0.17 & 34.48 & 0.77 \\
Fine-tune     & SYN+$L^{+}_{250}$ & 52.35  & 35.17 & 0.13 & 17.30 & 0.43 \\
SDA           & SYN+$L^{+}_{250}$ & 49.96  & 45.92 & 0.12 & 17.39 & 0.42 \\
LIRR-SDA (Ours) & SYN+$L^{+}_{250}$& 36.77 & 51.52 & 0.09 & 11.57 & 0.29 \\
\midrule
\hline
ORACLE        & $L^{+}_{500}$      & 89.41  & 16.71 & 0.14 & 29.48 & 0.65 \\
Fine-tune     & SYN+$L^{+}_{500}$  & 40.55  & 42.34 & 0.10 & 13.05 & 0.33 \\
SDA           & SYN+$L^{+}_{500}$  & 35.83  & 52.80 & 0.09 & 11.30 & 0.28 \\
LIRR-SDA (Ours) & SYN+$L^{+}_{500}$ & \textbf{32.36} & \textbf{54.51} & \textbf{0.08} & \textbf{9.93}  & \textbf{0.25} \\
\bottomrule
\hline
\end{tabular}
}
\end{table*}

\subsection{Evaluation Metrics}
We evaluated both keypoint regression and pose estimation accuracy using four metrics.

\begin{itemize}
    \item \textbf{Keypoint Error (KError)}: The mean Euclidean distance (in pixels) between the predicted 2D keypoints and the ground-truth 2D keypoints, averaged over all keypoints and test images.
    \item \textbf{Percentage of Correct Keypoints (PCK)}: The percentage of predicted keypoints whose Euclidean distance from the ground truth position is below a fixed pixel threshold (set according to the SPEED+ benchmark protocol).
    \item \textbf{Range Error}: The absolute error in estimated target range, expressed as a percentage of the true range.
    \item \textbf{Attitude Error}: The angular difference (in degrees) between the estimated and ground-truth spacecraft orientations, computed from their rotation matrices.
    \item \textbf{Pose Error}: The weighted metric defined in the SPEED+ evaluation protocol, which combines normalized translation error and rotation error into a single performance metric (lower is better).
\end{itemize}

\begin{figure*}[t]
    \centering
    \includegraphics[width=0.85\textwidth]{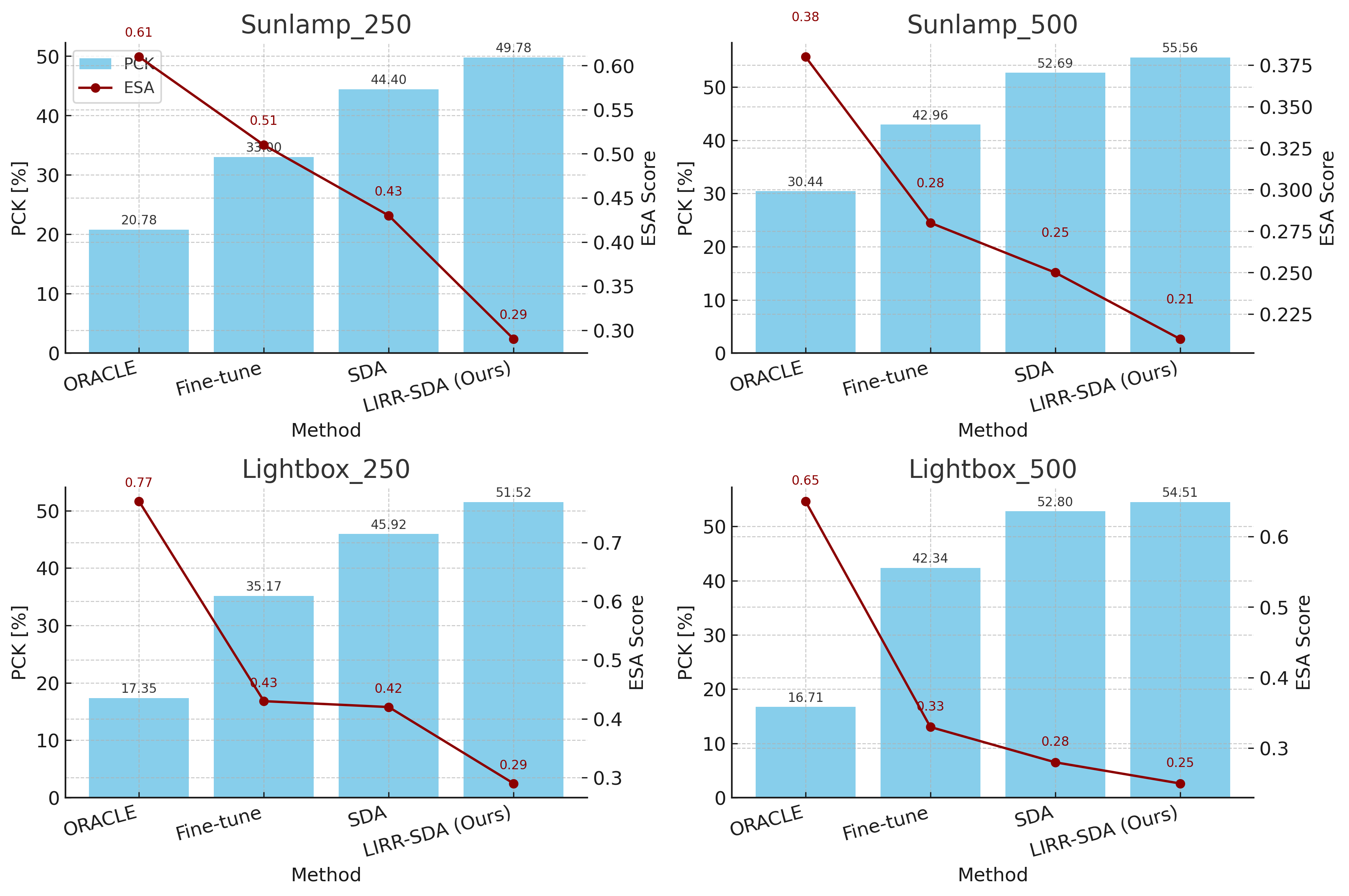}
    \caption{Comparison of PCK (bar, higher is better) and ESA score (line, lower is better) across different methods on the SPEED+ benchmark. Results are shown for (a) Sunlamp\_250, (b) Sunlamp\_500, (c) Lightbox\_250, and (d) Lightbox\_500 datasets. Our proposed LIRR-SDA consistently outperforms SDA, fine-tuning, and oracle baselines.}
    \label{fig:pck_esa_results}
\end{figure*}

\subsection{Quantitative Analysis}

Table~\ref{tab:quant_results} reports the performance of different training settings on the Sunlamp and Lightbox target domains. As expected, the \textbf{SOURCE-ONLY} baseline performs the worst, with very high keypoint errors and low PCK scores, reflecting the severity of the synthetic-to-real gap. The \textbf{ORACLE} setting improves results by using labeled target data but is constrained by limited annotations. \textbf{Fine-tuning} synthetic-pretrained models with target labels provides stronger results than Oracle but remains inconsistent, often underperforming compared to SDA.  

\textbf{SDA} consistently improves over both Oracle and Fine-tune by leveraging labeled source and target data jointly, reaching 52.69\% PCK on Sunlamp\_500 and 52.80\% on Lightbox\_500. The comparative trends are further illustrated in Figure~\ref{fig:pck_esa_results}, which plots PCK (bars) and ESA scores (lines) across all datasets. The figure highlights the consistent advantage of our proposed LIRR-SDA method over SDA, fine-tuning, and oracle baselines. Notably, the improvements are most pronounced on the Sunlamp\_500 and Lightbox\_500 settings, where larger labeled subsets allow LIRR-SDA to achieve both higher keypoint accuracy and lower ESA scores, further narrowing the synthetic-to-real gap.  

Our proposed \textbf{LIRR-SDA} achieves the best overall performance, obtaining the lowest ESA scores of 0.21 and 0.25 on Sunlamp\_500 and Lightbox\_500, respectively. These improvements confirm that jointly enforcing invariant representation learning and risk minimization leads to more accurate and reliable pose estimation under domain shift.

\begin{figure*}[t]
    \centering
    \includegraphics[width=\textwidth]{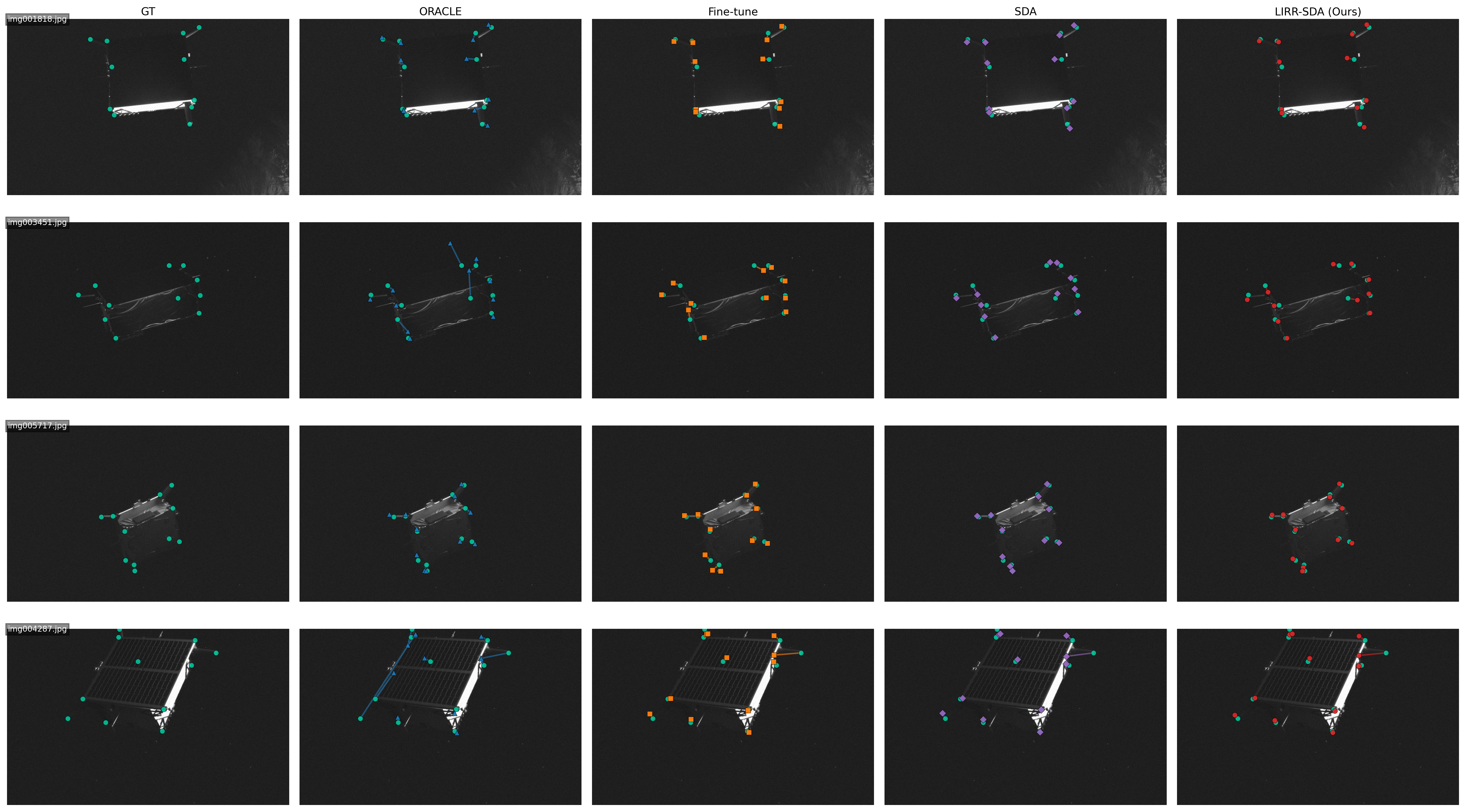}
    \caption{Qualitative comparison of keypoint regression on Lightbox images. LIRR-SDA achieves the closest alignment to ground truth (green) compared to ORACLE, Fine-tune, and SDA baselines.}
    \label{fig:qualitative_rows}
\end{figure*}

\subsection{Qualitative Analysis}
\label{subsec:qual_analysis}

Figure~\ref{fig:qualitative_rows} presents a qualitative comparison of keypoint regression results across representative target-domain test images. Ground truth keypoints are shown in green, while predictions from ORACLE, Fine-tune, SDA, and our proposed LIRR-SDA are overlaid in distinct colors. Lines connecting predicted and ground truth locations indicate the localization error for each keypoint.  

The results highlight several important trends consistent with the quantitative evaluation. First, the \textbf{ORACLE} baseline, trained only on a small number of labeled target samples, often produces large deviations from ground truth, particularly under challenging viewpoints or illumination conditions. \textbf{Fine-tuning} synthetic-pretrained models with target labels improves the predictions but frequently leads to misaligned or collapsed keypoints, suggesting overfitting to the limited labeled data. In contrast, \textbf{SDA} achieves better overall alignment with ground truth, but residual errors are still evident on difficult keypoints such as the antenna tips or occluded corners.  

Our proposed \textbf{LIRR-SDA} demonstrates the most consistent and accurate predictions across all examples. Keypoints remain tightly clustered around their ground truth locations, and error lines are noticeably shorter compared to competing methods. This visual evidence reinforces the effectiveness of jointly optimizing invariant representations and risk minimization, leading to improved keypoint localization even under significant synthetic-to-real domain shift.

\section{Conclusion}

This paper presented a supervised domain adaptation framework for 6-DoF spacecraft pose estimation, focusing on the keypoint regression stage of the hybrid pipeline. By integrating the LIRR principle into a fully supervised setting, we demonstrated that aligning invariant representations with task-specific risk enables effective use of both synthetic and scarce labeled real data.  
Our experiments on the SPEED+ benchmark confirm that this strategy minimizes the synthetic-to-real gap and consistently improves performance over conventional baselines. The gains are evident not only in terms of keypoint accuracy (PCK) but also in downstream pose estimation quality as measured by the ESA score. These results emphasize the importance of carefully designed adaptation objectives when labeled target data is limited but not entirely absent.
While the framework is lightweight and adaptable to different backbones, several challenges remain open. In particular, robustness under extreme illumination changes, occlusions, and motion blur is not fully resolved. Future work will explore temporal modeling across image sequences, cross-sensor adaptation, and reducing annotation requirements further through self- or weakly supervised learning.
In summary, the study highlights supervised domain adaptation as a promising direction for reliable spacecraft pose estimation, providing both practical insights for deployment and a foundation for further research into generalizable on-orbit perception.

\section*{Acknowledgments}
This research has been conducted in the context of the \texttt{DIOSSA} project, supported by the European Space Agency (ESA) under contract no. \texttt{[4000 1 4 4941241 NL/KK/adu]}. The simulations were performed on the Luxembourg National Supercomputer MeluXina. The authors gratefully acknowledge the LuxProvide teams for their expert support and the partners of the consortium for their contributions.


\end{document}